\documentclass[preprint,12pt]{elsarticle}

\usepackage{amssymb}
\usepackage{amsmath}

\usepackage[utf8]{inputenc} 
\usepackage[T1]{fontenc}    
\usepackage{hyperref}       
\usepackage{url}            
\usepackage{booktabs}       
\usepackage{amsfonts}       
\usepackage{nicefrac}       
\usepackage{microtype}      
\usepackage{xcolor}         

\usepackage{algorithm}
\usepackage{algorithmic}
\usepackage{multirow}
\usepackage{multicol}

\usepackage{microtype}
\usepackage{graphicx}
\usepackage{subfigure}
\usepackage{booktabs} 

\usepackage{amsmath}
\usepackage{amssymb}
\usepackage{mathtools}
\usepackage{amsthm}
\usepackage{MnSymbol}

\usepackage{inconsolata}

\usepackage{listings}
\journal{Knowledge-Based Systems}

\begin{document}

\begin{frontmatter}



\title{Visualizing, Rethinking, and Mining the Loss Landscape of Deep Neural Networks}

\author{Yichu Xu\corref{cor1}\fnref{label1,label2,cor2}}
\ead{xuyc@lamda.nju.edu.cn}
\cortext[cor1]{Corresponding author}
\author{Xin-Chun Li\fnref{cor2}\fnref{label1,label2, cor2}}
\ead{lixc@lamda.nju.edu.cn}
\fntext[cor2]{Equal contribution}
\author{Lan Li\fnref{label1,label2}}
\ead{lil@lamda.nju.edu.cn}

\author{De-Chuan Zhan\fnref{label1,label2}}
\ead{zhandc@nju.edu.cn}

\affiliation[label1]{organization={School of Artificial Intelligence, Nanjing University},
            city={Nanjing},
            country={China}}
\affiliation[label2]{organization={National Key Laboratory for Novel Software Technology, Nanjing University},
            city={Nanjing},
            country={China}}

\begin{abstract}
The loss landscape of deep neural networks (DNNs) is commonly considered complex and wildly fluctuated. However, an interesting observation is that the loss surfaces plotted along Gaussian noise directions are almost v-basin ones with the perturbed model lying on the basin. This motivates us to rethink whether the 1D or 2D subspace could cover more complex local geometry structures, and how to mine the corresponding perturbation directions. This paper systematically and gradually categorizes the 1D curves from simple to complex, including v-basin, v-side, w-basin, w-peak, and vvv-basin curves. Notably, the latter two types are already hard to obtain via the intuitive construction of specific perturbation directions, and we need to propose proper mining algorithms to plot the corresponding 1D curves. Combining these 1D directions, various types of 2D surfaces are visualized such as the saddle surfaces and the bottom of a bottle of wine that are only shown by demo functions in previous works. Finally, we propose theoretical insights from the lens of the Hessian matrix to explain the observed several interesting phenomena.
\end{abstract}




\end{frontmatter}


\section{Introduction}
It is commonly recognized that deep neural networks (DNNs) are difficult to train without the proposal of proper architectures ~\cite{bn, ResNet}, proper initialization~\cite{UnderstandDifficulty, ReLU, Kaiming}, or effective optimization algorithms~\cite{GradientOverview, AdaDelta, Adam}. A guess is that the loss landscape of DNNs is too complex to search for a qualified solution~\cite{UnderstandDifficulty, Goldilock, NoBadValley, VisualizingLandscape}. Capturing a global view of the high-dimensional loss landscape of DNNs is still a mystery to the community~\cite{Intrinsic, LargeScaleLandscape, Emergent, LandscapeImplicitBias, GlobalView, WideSurface, GeneralizeVis}, but projecting it into the 1D or 2D subspace is a common way to visualize the local geometry of DNNs~\cite{Goodfellow,im2016empirical, VisualizingLandscape, VisualBERT, DeepEnsembleLandscape, Taxonomize}.

Previous works either show 1D or 2D surfaces that are nearly smooth and are not as complex as expected~\cite{Goodfellow, PyHessian, im2016empirical, FGE}, or they only illustrate the complex DNN loss landscape through simple demo functions~\cite{Saddle, TheRole, Sharp}. For example, the linear interpolation between the initialization and the converged solution shows monotonic decreasing losses, i.e., the Monotonic Linear Interpolation (MLI) phenomenon~\cite{Goodfellow, RevisistMLI, AnalyzingMLI, WhatCan, PlateauMLI}; the linear interpolation between two independent solutions will commonly encounter one and only one loss barrier, and they are amazingly connected by simple Bezier or quadratic curves, i.e., the Linear Mode Connectivity (LMC) phenomenon~\cite{Goodfellow, FGE, OTFusion, GitReBasin}. The existence of MLI and LMC makes us rethink the loss landscape of DNNs: {\it could the 1D curves or 2D surfaces display more complex patterns? If so, could we search for definite ways to mine and visualize them in an explainable manner?}

\begin{figure}[tbp]
	\centering
	\includegraphics[width=\linewidth]{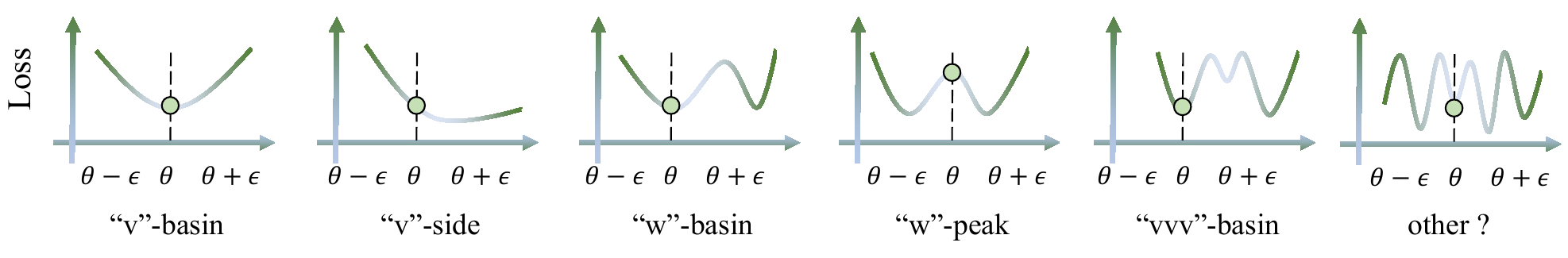}
	\caption{The illustration of the categorized 1D curves. For each categorized type, we provide definite perturbation directions or mining algorithms to plot them in an explainable way. An open question is whether we can mine more perturbation directions and plot complex 1D curves correspondingly.} \label{fig:teaser}
\end{figure}

This paper systematically and gradually visualizes the loss surface of mainstream DNNs in 1D or 2D subspace. We first find an interesting observation that perturbing DNN parameters by Gaussian noise directions commonly leads to monotonic increasing curves on both two sides. These types of curves are categorized as v-basin curves which look like the shape of ``v'' and the perturbed model lies on the basin. Setting the perturbation direction to the negative gradient or the direction to subsequent checkpoints could display v-side curves. The w-basin curves are inspired by the loss barrier phenomenon in LMC~\cite{Goodfellow, OptimizeMC, NoBarrier, FGE}, where we take the direction to an independently trained checkpoint as the perturbation direction. Plotting w-peak curves should dive into the Hessian eigenvalues and eigenvectors of the DNN~\cite{Saddle, EigenvaluesBeyond, EmpiricalHessian, GradSubspace, NegativeHessian, Sharpest, HessianDensity, HessianEigenspectrum}. However, the most negative eigenvalue of the Hessian is absolutely small, which makes the w-peak curves own small curvatures and only show a short loss decreasing trend on both sides. We propose an algorithm to mine obvious w-peak curves and analyze the relationship to the Hessian eigenvector directions. There are no intuitive perturbation directions to plot more complex curves such as vvv-basin ones, and we also provide an algorithm to mine them. The illustration can be found in Fig.~\ref{fig:teaser}. 

Combining the above 1D directions, we could plot various types of 2D surfaces such as saddle surfaces and a gutter structure like the bottom of a bottle. Notably, these surfaces are seldom plotted in previous works, which are instead illustrated by simple demo functions such as in~\cite{Saddle}. Aside from empirical visualization, this paper additionally provides theoretical insights to explain the phenomenon of monotonic loss increasing when perturbed by Gaussian noise. Furthermore, this explanation is bridged to the MLI phenomenon, which leads to a novel insight for explaining MLI.

To conclude, this paper has several advantages and contributions as follows: (1) systematically visualizing and mining the embedded 1D loss curves of DNNs by category; (2) plotting several types of 2D surfaces which are only illustrated by demo functions in previous works; (3) defining as GMI the monotonic loss increasing phenomenon when perturbed by Gaussian noise and providing theoretical insights from the lens of Hessian for GMI and MLI; (4) making it easier to understand and interpret the loss surfaces of DNNs in a step-by-step manner; (5) leaving open problems and interesting guesses about the low-dimensional loss landscape visualization of DNNs.

\section{Empirical Visualization of Loss Landscape} \label{sec:empirical}
This section introduces basic notations and experimental settings, then visualizes and mines 1D curves by category, and finally presents various types of 2D surfaces.

\subsection{Basic Notations and Experimental Settings} \label{sec:notation}
Given a model $\theta$, we could perturb it along a single direction $\epsilon$ or two orthogonal directions $\epsilon_1$ and $\epsilon_2$. The former shows the 1D curve embedded in the global landscape, while the latter plots the 2D surface centered around $\theta$. We uniformly name $\epsilon$, $\epsilon_1$, and $\epsilon_2$ perturbation directions or noise directions, and name $\theta$ the perturbed model. $\epsilon$ has the same shape as $\theta$, and is usually element-wisely sampled for every trainable parameter. By default, the 1D curves are plotted in the range of $\lambda \in [-1, 1]$ with the perturbation equation as $\theta + \lambda \epsilon$. We use the equation of $\theta + \lambda_1 \epsilon_1 + \lambda_2 \epsilon_2$ to plot the 2D surfaces with $\lambda_1 \in [-1, 1]$ and $\lambda_2 \in [-1, 1]$. Sometimes, we set the range of $\lambda$, $\lambda_1$, or $\lambda_2$ as $[-s, s]$ to show a micro or macro landscape view. The norm of the perturbation direction may also be scaled to the same as the perturbed model by the equation $\epsilon \leftarrow ||\theta|| * \epsilon / ||\epsilon||$. 

We train an MLP with two layers on CIFAR-10 (C10)~\cite{cifar}, and train ResNet32/110 (RN32/110)~\cite{ResNet} on CIFAR-100 (C100)~\cite{cifar}. We also finetune the pre-trained MobileNet-V2 (MV2)~\cite{MobileNetV2} on CUB~\cite{CUB}. The pre-trained ResNeXt101 (RNX101)~\cite{ResNeXt} could be directly downloaded and verified on ImageNet~\cite{ImageNet}. Pre-trained models are from \texttt{PyTorch}~\footnote{\url{https://pytorch.org/}}. We do not consider perturbing the running statistics in the BN layers~\cite{bn}, and we will forward the interpolated model one pass on the dataset to re-calculate them. More experimental details can be found in Appendix~\ref{supp-sec:exper-detail}.

\subsection{Visualization of v-basin Curves} \label{sec:v-basin}
\begin{figure}[tb]
	\centering
	\includegraphics[width=\linewidth]{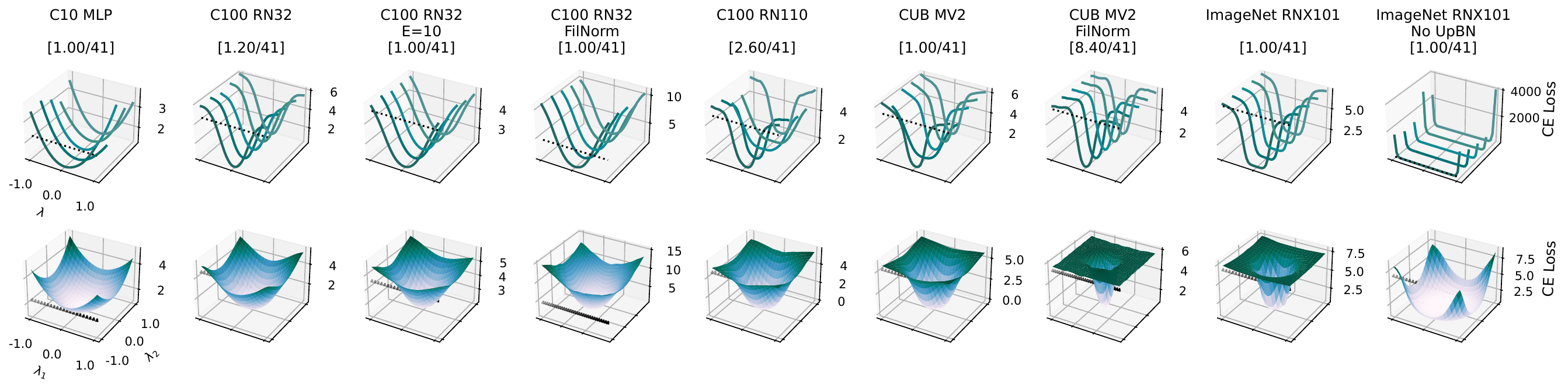}
	\caption{The 1D v-basin curves and 2D surfaces along Gaussian noise directions. The plots are amazingly smoother than we previously thought under various conditions.} \label{fig:gaussian}
\end{figure}

A common way to view the local landscape of DNNs is setting $\epsilon$, $\epsilon_1$, and $\epsilon_2$ as random Gaussian directions. The 1D curves and 2D surfaces are displayed in Fig.~\ref{fig:gaussian}. The horizontal dotted black lines represent the random prediction loss threshold, i.e., the loss of $\log(C)$ with $C$ being the number of classes. Various conditions are considered, which include: (1) the number of training epochs (E), where we show the results of RN32 on C100 when $E=10$, and the others are $E=200$ by default; (2) the filter normalization (``FilNorm'') method proposed by~\cite{VisualizingLandscape}, which normalizes each filter in the noise direction to have the same norm of corresponding filters in the perturbed model; (3) the processing of BN statistics, and ``No UpBN'' denotes that the BN running states are not re-calculated. Five independent 1D curves are visualized in the same plot. The average number of stationary points in the five curves is reported in the ``[]''. We plot 41 discrete points for each 1D curve, and most of them only have 1 stationary point on average (i.e., the perturbed model itself when $\lambda=0.0$). One exception is the 7th plot that utilizes FilNorm on CUB, while the additional stationary points almost exist in line segments that surpass the random prediction loss. We denote these types of 1D curves as v-basin curves with the perturbed model lying on the basin. 

\begin{figure}[tb]
	\centering
	\includegraphics[width=\linewidth]{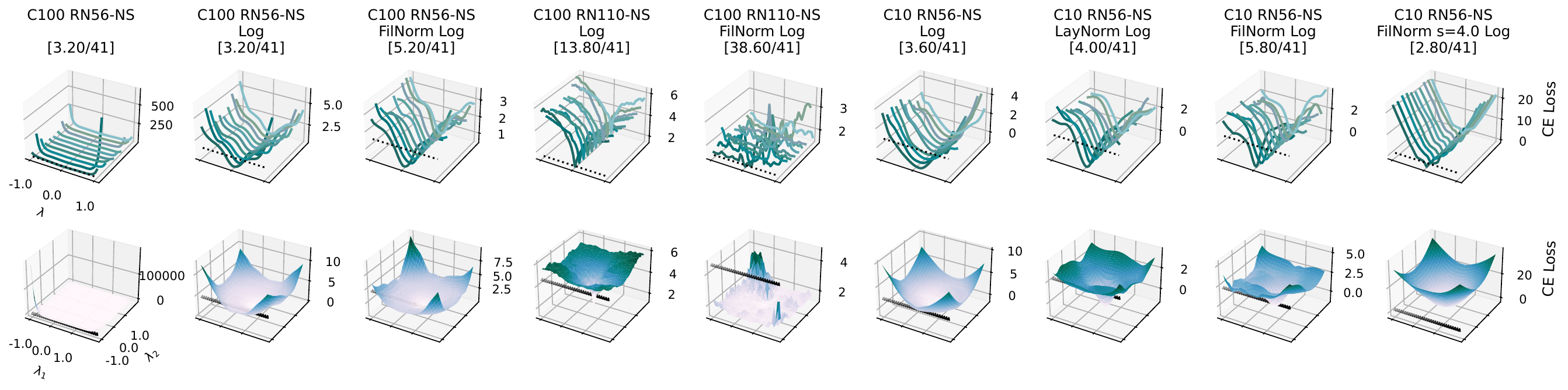}
	\caption{The 1D v-basin curves and 2D surfaces plotted around models with no skip connection along Gaussian noise directions. ``Log'' denotes that the y-axis is in the log scale.} \label{fig:gaussian-noskip}
\end{figure}

The previous work~\cite{VisualizingLandscape} plots the 2D surface of ResNet without skip connections (RN56/110-NS), which is extremely chaotic. We carefully checked their code and found several interesting phenomena. First, perturbing RN-NS could easily lead to exploded losses (e.g., a loss value of $10^5$ on C100 using RN56-NS), which makes the landscape look like a horizontal hyperplane (i.e., the first column in Fig.~\ref{fig:gaussian-noskip}). As a trick, they plot the losses in a log scale to enhance the distinctness between points. Second, RN110-NS is hard to optimize, which only performs slightly better than random guess (i.e., a test accuracy of $7.08\%$ on C100 and $10.09\%$ on C10). Perturbing models that are not well-trained is less meaningful, and the loss landscape around them is chaotic (e.g., the 5th column in Fig.~\ref{fig:gaussian-noskip}). Third, RN56-NS could achieve much better performances on C100 ($57.06\%$) and C10 ($87.01\%$), and the chaotic loss landscape almost exists above the random guess curves (i.e., above the dotted black lines). Additionally, with a macro view scale (i.e., $s=4.0$), the loss landscape of RN56-NS on C10 becomes smooth again (i.e., the last column in Fig.~\ref{fig:gaussian-noskip}). With the same number of scatter points, a macro view means a larger step between adjacent scatters, which may skip the fluctuating segments. The theoretical insights will be provided in Sect.~\ref{sec:theory}.

Overall, chaotic surfaces seem to show up only on poorly trained DNNs or in regions worse than random predictions, which are of little significance to study. For well-trained and mainstream models, the loss surfaces along Gaussian perturbations are extremely smoother than we previously thought. In other words, it seems difficult to plot 1D curves other than the v-basin ones by Gaussian perturbation alone. We denote the phenomenon that Gaussian perturbation leads to double-side monotonic increasing losses as Gaussian Monotonic Increasing (GMI). A theoretical parallel will be bridged between GMI and MLI in Sect.~\ref{sec:theory}. Some experimental details are explained in Appendix~\ref{supp-sec:fig}.


\subsection{Visualization of v-side Curves} \label{sec:v-side}
\begin{figure}[tb]
	\centering
	\includegraphics[width=\linewidth]{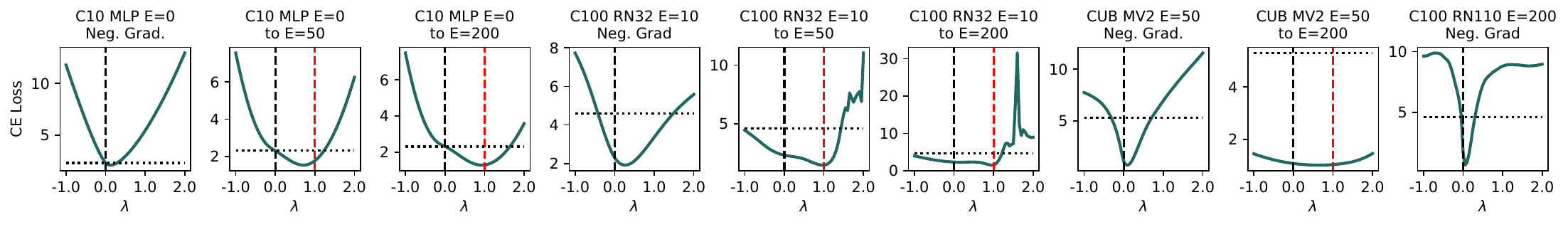}
	\caption{The 1D v-side curves plotted around different checkpoints. The vertical black/red dotted line shows the position of the perturbed/subsequent checkpoint.} \label{fig:descent-curves}
\end{figure}

Considering the first-order and second-order approximation (f.o.a and s.o.a) of the loss $\mathcal{L}(\theta + \lambda\epsilon)$:
\begin{equation}
    \mathcal{L}_{\text{f.o.a}} \approx \mathcal{L}(\theta) + \lambda \epsilon^T g_{\theta}, \,\,\,\,
    \mathcal{L}_{\text{s.o.a}} \approx \mathcal{L}(\theta) + \lambda \epsilon^T g_{\theta} + \frac{1}{2}\lambda^2 \epsilon^T H_{\theta} \epsilon, \label{eq:approx}
\end{equation}
where $g_{\theta}=\nabla_{\theta} \mathcal{L}$ and $H_{\theta}=\nabla\nabla^T \mathcal{L}$ denote the gradient and Hessian matrix calculated at the point of $\theta$. The first-order approximation implies that the negative gradient is the sharpest descent direction, i.e., $\epsilon=-g_{\theta}$. However, the gradient norm of converged models is nearly zero, only showing negligible descent losses. Hence, we consider the intermediate checkpoints and take the negative gradient as one type of descent direction. The direction to the subsequent checkpoint (e.g., $\epsilon = \theta_{E=200} - \theta_{E=50}$) is also a descent direction for $\theta_{E=50}$, which is inherently an accumulation of multiple gradient steps. Fig.~\ref{fig:descent-curves} shows the plotted results, where we take MLP with $E=0$ on C10, RN32 with $E=10$ on C100, MV2 with $E=50$ on CUB, and RN110 with $E=200$ on C100. The curves are almost v-side ones with the perturbed model lying on the valley side (i.e., the vertical black dotted line when $\lambda=0.0$). RN32 on C100 again shows chaotic segments whose loss surpasses that of a random guess. 
 
\begin{figure}[tb]
	\centering
	\includegraphics[width=\linewidth]{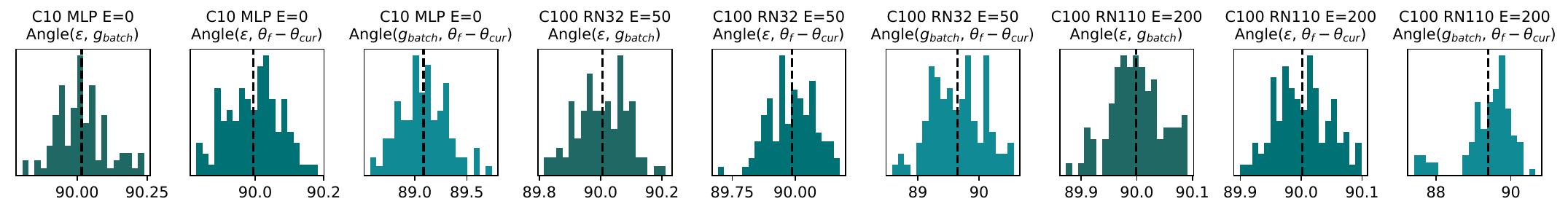}
	\caption{The angles between randomly sampled Gaussian directions and the descent directions. $g_{\text{batch}}$ denotes the negative of batch gradient, and $\theta_{\text{cur}}$/$\theta_\text{f}$ denotes the current/final checkpoint.} \label{fig:descent-angles}
\end{figure}

According to the first-order approximation, the loss will descend in a specific range if we could sample a Gaussian vector satisfying $\epsilon^T g_{\theta} < 0$. Easily, the mean and variance of $\epsilon^T g_{\theta}$ is $\mathbb{E}_{\epsilon}[\epsilon^T g_{\theta}]=0$ and $\mathbb{V}_{\epsilon}[\epsilon^T g_{\theta}]=||g_{\theta}||_2^2$. The probability of sampling a same vector as $g_{\theta}$ is $\exp(-\frac{1}{2}||g_{\theta}||_2^2)/(2\pi)^{d/2}$, and $d$ is the total number of trainable parameters. Take the MLP on C10 as an example, it has about d=395K parameters, and $||g_{\theta}||_2^2$ is about $119.7$ under $E=0$, and the logarithmic probability of sampling $g_{\theta}$ is about $-3.6\times 10^5$. That is, we could hardly sample a Gaussian direction having a larger overlap with the gradient direction in the high-dimensional space. We sample 100 groups of Gaussian directions and calculate their angles with the negative gradient direction and the direction to the subsequent checkpoint. Fig.~\ref{fig:descent-angles} shows the distribution of angles. The gradient directions are calculated on random data batches. The angle range between the Gaussian direction and descent directions is about $[89.75, 90.25]$. For comparison, we also calculate the angle between the negative gradient and the direction of the converged model, which are almost in the range of $[88.0, 90.0]$. This shows that the one-step batch gradient may slightly imply the global converged direction, while it is nearly impossible to sample a Gaussian noise that has an acceptable overlap with descent directions, e.g., a noise with an angle smaller than 89.5.

\begin{figure}[tb]
	\centering
	\includegraphics[width=\linewidth]{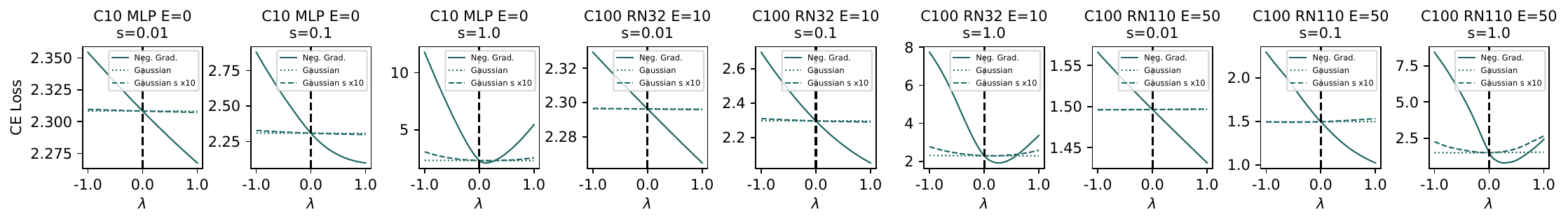}
	\caption{The descent curves of the negative gradient and the most overlapped Gaussian noise.} \label{fig:descent-steps}
\end{figure}

We select the Gaussian noise most overlapped with the negative gradient direction from the 100 groups, and plot the descent curves in Fig.~\ref{fig:descent-steps}. We set $s$ in $\{0.01, 0.1, 1.0\}$ and scale the norm of the Gaussian direction to that of the negative gradient. In each plot, the x-axis with Gaussian noise is also scaled by $10\times$ for better comparison. Even with a $10\times$ scale, the loss decreasing brought by the Gaussian direction is negligible. This verifies the GMI phenomenon shown in Fig.~\ref{fig:gaussian} from a micro perspective that the Gaussian noise could hardly lead to v-side curves.

\subsection{Visualization of w-basin/w-peak Curves} \label{sec:w-basin-peak}
\begin{figure}[tbp]
	\centering
	\includegraphics[width=\linewidth]{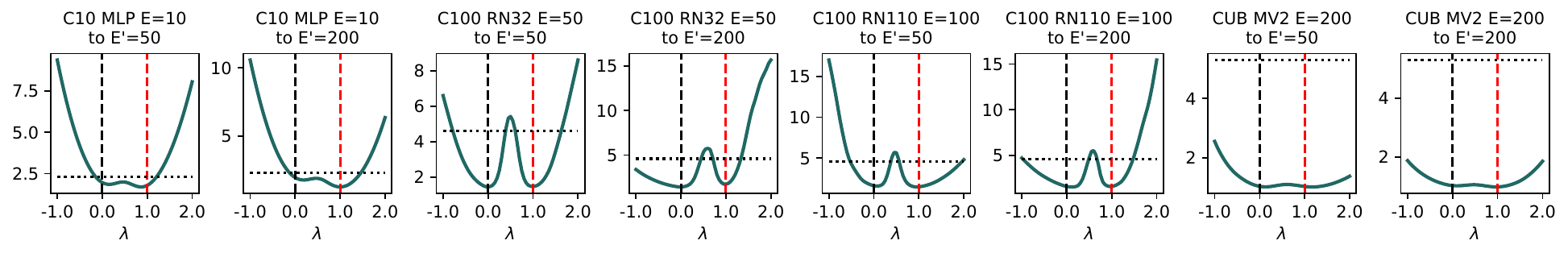}
	\caption{The 1D w-basin curves inspired by loss barrier between independently trained checkpoints. The vertical black/red dotted lines show the positions of two independent checkpoints.} \label{fig:w-curves}
\end{figure}

Previous researches about the mode connectivity of DNNs provide an intuitive way to plot w-basin curves~\cite{Goodfellow, NoBarrier, ModeVolume}. Commonly, the linear interpolation between two independent converged models encounters one and only one barrier. That is, the loss $\mathcal{L}((1-\lambda)\theta_1 + \lambda \theta_2)$ within $\lambda \in [0, 1]$ looks like a hill. In other words, given a model $\theta_1$, we could plot the 1D curve by the equation of $\theta_1 + \lambda(\theta_2 - \theta_1)$. If we take $\lambda \in [-1, 2]$, then we could obtain w-basin curves as shown in Fig.~\ref{fig:w-curves}. Notably, we extend the phenomenon of loss barrier to different checkpoints. For example, the vertical black dotted line denotes the position of the MLP checkpoint with $E=10$, and the red dotted line denotes another independent checkpoint with $E^\prime=50$ (i.e., the 1st plot in Fig.~\ref{fig:w-curves}). All plots present perfectly smooth w-basin curves, showing no chaotic segments. 

Then, we would like to search for a direction that could make the loss double-side decrease, i.e., w-peak curves. If we assume $g_{\theta}$ is zero for a converged moel $\theta$, then the second-order approximation in Eq.~\ref{eq:approx} will become $\mathcal{L}_{\text{s.o.a}} - \mathcal{L}(\theta) \approx \frac{1}{2}\lambda^2 \epsilon^T H_{\theta} \epsilon$. Setting $\epsilon$ to the eigenvectors of the Hessian matrix corresponding to the negative eigenvalues (abbreviated as N.E.) could make $\mathcal{L}_{\text{s.o.a}}$ smaller than $\mathcal{L}(\theta)$ within a definite small range of $\lambda$. We use the \texttt{sparse.linalg.eigsh} in the \texttt{Scipy}~\footnote{\url{https://docs.scipy.org/doc/scipy/reference/generated/scipy.sparse.linalg.eigsh.html}} package to calculate the most 5 positive or negative eigenpairs~\cite{NegativeHessian}. Specifically, \texttt{eigsh} accepts as input the \texttt{LinearOperator} constructed by the ``Jacobian Vector Product'' function that returns $H_{\theta}v$ for a given vector $v$. The detail and demo code can be found in Appendix~\ref{supp-sec:demo-eigsh}. Fig.~\ref{fig:hessian-1d} plots the 1D curves along the positive eigenvectors (P.E.) and N.E. ones. We show results of MLP on C10, RN32 on C100, and MV2 on CUB, and each pair considers the checkpoint of $E \in \{10, 50, 200\}$. The 1D curves along P.E. directions almost consistently show v-basin curves, while the cases for N.E. ones are complex. The initial checkpoints of simple models (e.g., MLP on C10 and RN32 on C100) display perfect w-peak curves when $E=10$ or $E=50$. MV2 on CUB does not display obvious w-peak curves. Additionally, with the model becoming converged, the ``height of the peak'' decreases.


\begin{figure}[tbp]
	\centering
	\includegraphics[width=\linewidth]{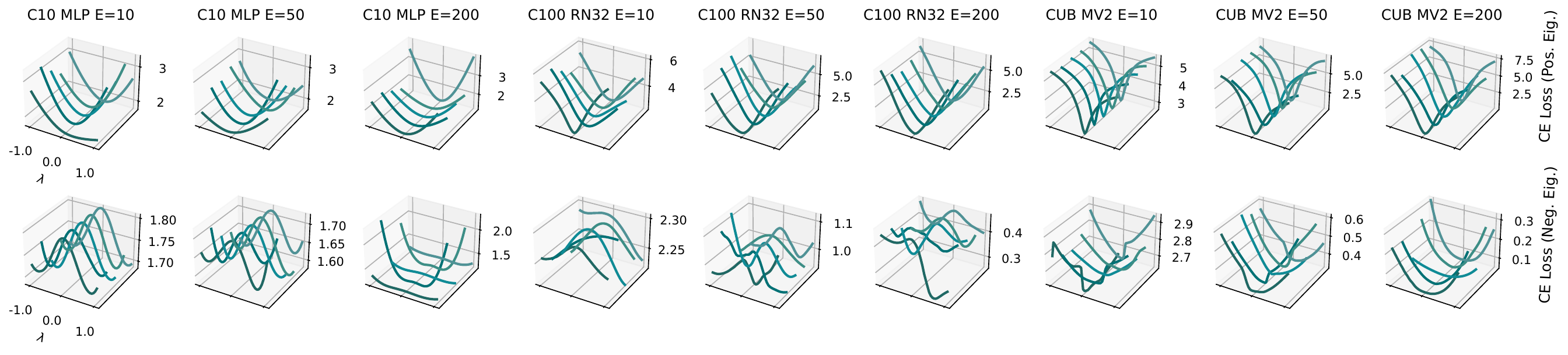}
	\caption{The 1D curves plotted along the eigenvector directions of the Hessian matrix. The second row shows the w-peak curves along N.E. directions. Top 5 directions of P.E. and N.E. are plotted.} \label{fig:hessian-1d}
\end{figure}

We also propose an algorithm to mine the w-peak curves without calculating the negative eigenvalues and corresponding eigenvectors of the Hessian matrix. Our optimization formulation is:
\begin{equation}
    \arg\min_{\epsilon} \mathbb{E}_{\lambda\in [-1, 1]}\left[\mathcal{L}(\theta + \lambda \epsilon)\right]. \label{eq:mine}
\end{equation}

\begin{figure}[tbp]
	\centering
	\includegraphics[width=\linewidth]{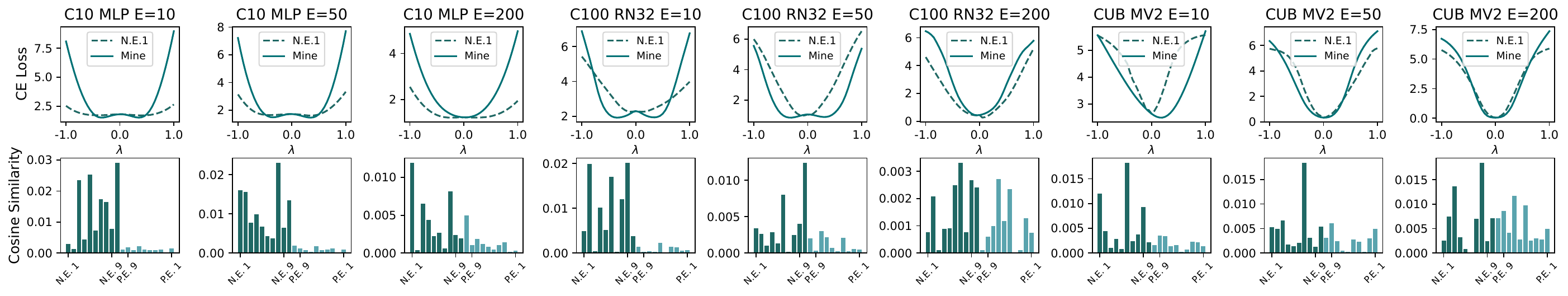}
	\caption{The mined w-peak curves via Algo.~\ref{algo:mine} and the cosine similarity with P.E./N.E. directions.} \label{fig:mine}
\end{figure}

To solve this optimization problem, we first initialize the elements in $\epsilon$ as zero. For each data batch, we sample $\lambda$ from $[-\alpha, \alpha]$ and obtain the interpolated model $\hat{\theta} = \theta + \lambda \epsilon$. The loss and gradient are calculated on $\hat{\theta}$ and we update $\epsilon$ by $\epsilon - \eta\lambda \nabla_{\hat{\theta}}\mathcal{L}$. $\eta$ is the learning rate and $\alpha$ will gradually increase from $0$ to $1$ during the whole optimization process. The pseudo-code is listed in Algo.~\ref{algo:mine}. With the optimized $\epsilon$, we plot the 1D curves of $\theta + \lambda \epsilon$ in the first row of Fig.~\ref{fig:mine}. Similarly, it is hard to obtain w-peak curves for MV2 fine-tuned on CUB. For MLP on C10 and RN32 on C100, the w-peak curves are obvious when $E$ is smaller. When $E=200$, it is hard to obtain double-side loss decreasing curves because the converged model already reaches a quite low-loss area. We also study the relationship between the mined direction and the eigenvectors of $H_{\theta}$. The cosine similarities are calculated and reported in the second row of Fig.~\ref{fig:mine}. The first $10$ bars show the absolute cosine similarity with ``N.E.1'' to ``N.E.10'' while the following $10$ bars show that with ``P.E.10'' to ``P.E.1''. The mined direction is not just one of the Hessian eigenvectors but seems to be a weighted combination of all eigenvectors with the weights of N.E. slightly larger than that of P.E. ones.

\begin{multicols}{2}
    \begin{algorithm}[H]
		\caption{{Mine w-peak Curves}}
		\label{algo:mine}
		\begin{algorithmic}[1]
			\FOR{each epoch $e = 1, 2, \ldots, E$}
            \STATE Set $\alpha=e/E$
			\FOR{each batch}
			\STATE Sample $\lambda \in [-\alpha, \alpha]$ uniformly
            \STATE Interpolate model $\hat{\theta}=\theta+\lambda\epsilon$
            \STATE $\epsilon \leftarrow \epsilon - \eta \lambda \nabla_{\hat{\theta}}\mathcal{L}$
			\ENDFOR
			\ENDFOR
		\end{algorithmic}
      \end{algorithm}
      \begin{algorithm}[H]
		\caption{{Mine vvv-basin Curves}}
		\label{algo:mineplus}
		\begin{algorithmic}[1]
			\FOR{each epoch $e = 1, 2, \ldots, E$}
			\FOR{each batch}
			\STATE Sample $\lambda\in[0.5-\alpha, 0.5+\alpha]$
            \STATE Interpolate $\hat{\phi}=(1-\lambda)\theta+\lambda\phi$
            \STATE $\phi \leftarrow \phi - \eta (\nabla_{\phi}\mathcal{L} + \gamma\lambda \nabla_{\hat{\phi}}\mathcal{L})$
			\ENDFOR
			\ENDFOR
            \STATE $\epsilon=\phi-\theta$
 		\end{algorithmic}
      \end{algorithm}
\end{multicols}

\begin{figure}[tbp]
	\centering
	\includegraphics[width=\linewidth]{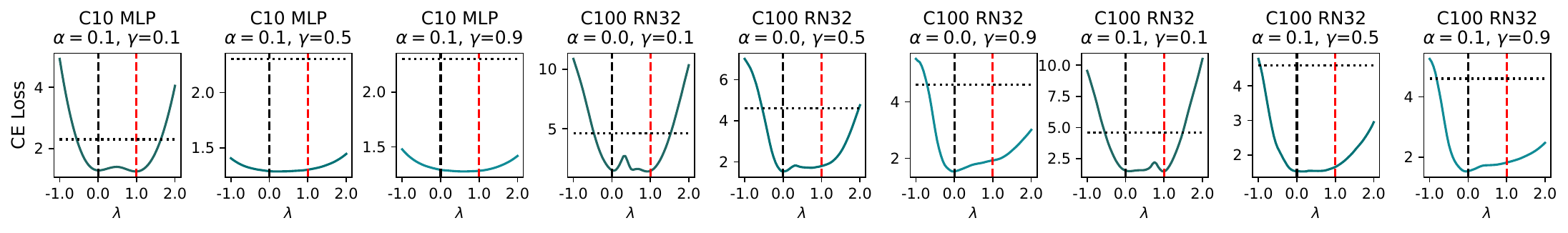}
	\caption{The mined vvv-basin curves via Algo.~\ref{algo:mineplus} under various hyperparameters of $\gamma$ and $\alpha$.} \label{fig:mineplus}
\end{figure}

\subsection{Visualization of vvv-basin Curves} \label{sec:vvv-basin}
We have not yet found an intuitive perturbation direction to plot vvv-basin curves, but we provide a possible optimization problem to mine possible directions: \begin{equation}
    \arg\min_{\phi} \mathbb{E}_{\lambda\in [0.5 - \alpha, 0.5 + \alpha]}\left[\mathcal{L}(\phi) + \gamma \mathcal{L}((1 - \lambda)\theta + \lambda \phi)\right], \label{eq:mineplus}
\end{equation}
where $\theta$ is a converged model, and $\gamma$ is the regularization coefficient, and $\alpha$ controls the interpolation range. This formula optimizes the loss of $\phi$ and simultaneously decreases the loss around the middle interpolation with $\theta$. The optimization process is similar to Eq.~\ref{eq:mine} and the pseudo-code is in Algo.~\ref{algo:mineplus}. When the optimization finishes, the direction of $\epsilon=\phi-\theta$ is utilized to plot 1D curves around $\theta$. We set $\alpha \in \{0.0, 0.1\}$ and $\gamma \in \{0.1, 0.5, 0.9\}$, respectively. The mined curves are shown in Fig.~\ref{fig:mineplus}, where the vertical black/red dotted lines mark the position of $\theta$/$\phi$. MLP on C10 is a simple DNN and shows no vvv-basin patterns. A proper hyperparameter group on C100 with RN32 may present vvv-basin curves (e.g., the 4th plot with $\alpha=0.0$ and $\gamma=0.1$). Other plots show patterns looking like the transitional shape from ``w'' to ``vvv''. Overall, vvv-basin curves have already become not so intuitive to mine and visualize.

\subsection{Visualization of 2D Surfaces by Combining 1D Directions} \label{sec:2d-surface}
\begin{figure}[tb]
	\centering
	\includegraphics[width=\linewidth]{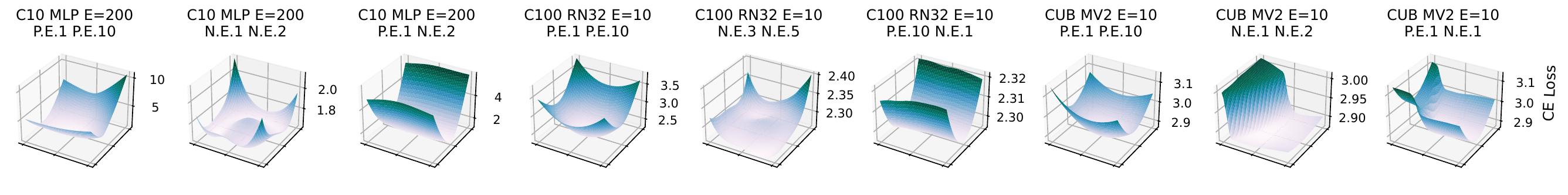}
	\caption{The 2D surfaces plotted by combining the eigenvector directions. P.E./N.E. denotes eigenvectors corresponding to the positive/negative eigenvalues.} \label{fig:hessian-2d}
\end{figure}

First, the Hessian eigenvector directions could be combined to plot the 2D surfaces. Eigenvectors are orthogonal to each other and we do not need to make them orthogonal again. The plots are in Fig.~\ref{fig:hessian-2d}. Combining two directions of P.E. leads to a surface looking like the 2D surfaces of Gaussian perturbation shown in Fig.~\ref{fig:gaussian}. Combining directions of N.E. shows a surface looking like the bottom of a bottle of wine, which is only illustrated by demo functions in previous works~\cite{Saddle}. An exceptional case is the 8th plot, which is resulted from the smaller negative curvature of MV2 on CUB shown in Fig.~\ref{fig:hessian-1d}. The combination of P.E. and N.E. leads to saddle surfaces, and the curvature along the N.E. direction is smaller because the absolute value of the corresponding eigenvalue is small (Fig.~\ref{fig:hessian-stats}). Then, the 1D directions that lead to v-basin, v-side, w-basin, w-peak, and vvv-basin curves could be combined to display more complex 2D surfaces. The possible 2D surfaces are shown in Fig.~\ref{fig:combine-2d}. The summary of the mined perturbation directions and their abbreviations are in Appendix~\ref{supp-sec:summary}.

\begin{figure}[tb]
	\centering
	\includegraphics[width=\linewidth]{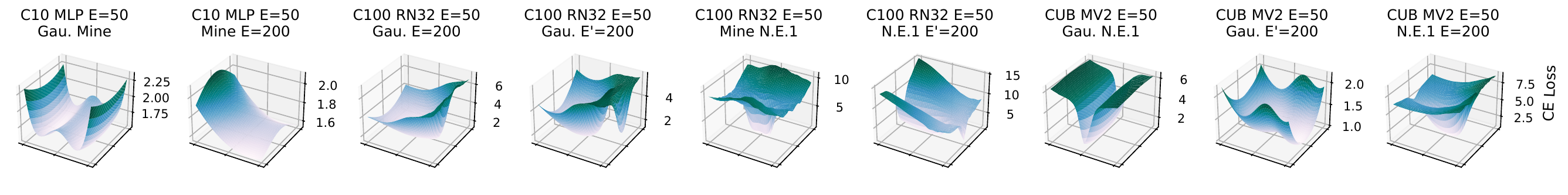}
	\caption{The 2D surfaces plotted by combining the mined 1D perturbation directions. The second line of the title shows the abbreviations of the utilized two directions.} \label{fig:combine-2d}
\end{figure}

\section{Theoretical Insights from the Lens of the Hessian} \label{sec:theory}
\begin{figure}[tb]
	\centering
	\includegraphics[width=\linewidth]{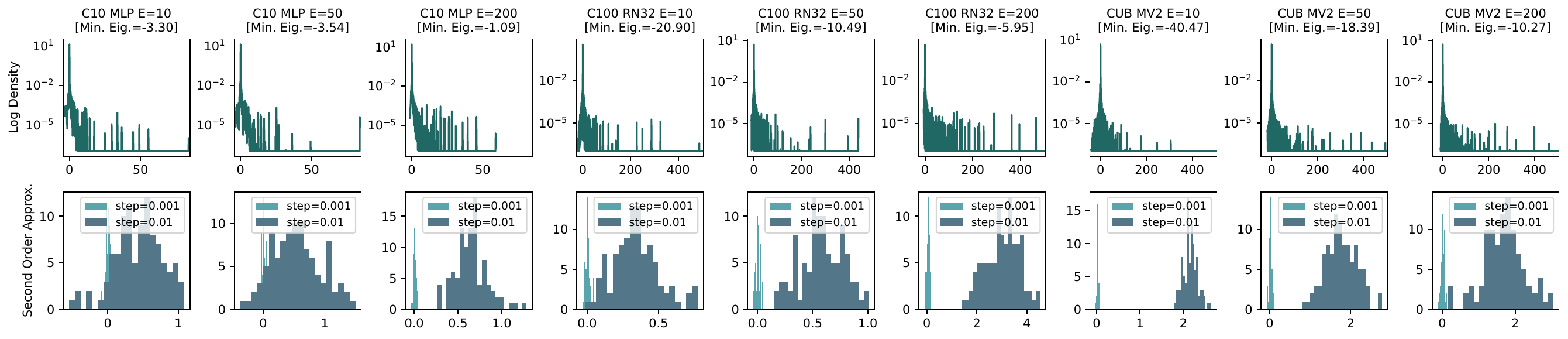}
	\caption{The eigenvalue density of the Hessian matrix (first row) and the distribution of loss change when perturbed by Gaussian noises with $\lambda \in \{0.001, 0.01\}$ (second row).} \label{fig:hessian-stats}
\end{figure}

This section provides some initial theoretical explanations for the observed interesting phenomena from the lens of the Hessian~\cite{HessianDensity, NegativeHessian, EmpiricalHessian}. Specifically, we utilize the \texttt{PyHessian}~\cite{PyHessian}\footnote{\url{https://github.com/amirgholami/PyHessian}} tool to calculate the approximated eigenvalue density of the Hessian matrix. The details and demo code are in Appendix~\ref{supp-sec:demo-pyhessian}. The logarithmic probability density of Hessian eigenvalues is shown in the first row of Fig.~\ref{fig:hessian-stats}. The Hessian density presents a shape with a bulk of values around zero and several large positive outliers~\cite{EmpiricalHessian, GradSubspace}. The smallest eigenvalue is reported in ``[]'', and its absolute value becomes smaller when $E$ increases. This means that the singularity of the Hessian matrix decreases and it is harder to mine w-peak curves when the model goes to a more convex area (Fig.~\ref{fig:hessian-1d} and Fig.~\ref{fig:mine}).


Then, we rethink the interpolation formula in MLI~\cite{Goodfellow, RevisistMLI, AnalyzingMLI, PlateauMLI} and the finding of GMI in our paper. The former plots the losses of $(1-\alpha)\theta_0 + \alpha \theta_f$ with $\alpha \in [0, 1]$, where $\theta_0$ and $\theta_f$ denote the initial and converged model, respectively. We could re-formulate this equation as $\theta_f + \lambda \epsilon$ with $\lambda = \alpha - 1 \in [-1, 0]$, and $\epsilon = \theta_f - \theta_0$. Hence, MLI and GMI differ only in the perturbation direction, which implies that they may have the same explanation. Considering the second-order approximation in Eq.~\ref{eq:approx}, the loss change when perturbed by the Gaussian noise is $\delta\mathcal{L} = \lambda\epsilon^Tg_{\theta} + \frac{1}{2}\lambda^2 \epsilon^T H_{\theta} \epsilon$. The mean and variance of this term is:
\begin{equation}
    \mathbb{E}_{\epsilon}\left[ \delta\mathcal{L} \right] = \frac{1}{2}\lambda^2 \sigma^2 tr(H_{\theta}), \,\, \mathbb{V}_{\epsilon}\left[ \delta\mathcal{L}\right] = \lambda ||g_{\theta}||_2^2 + \frac{1}{2}\lambda^2 \sigma^4 tr(H_{\delta}H_{\delta}), \label{eq:mean-var}
\end{equation}
where we assume $\epsilon$ is element-wisely sampled from $\mathcal{N}(0, \sigma^2)$, and $tr(\cdot)$ denotes the trace of a matrix. The trace of the Hessian during the training process is almost always positive as presented in previous studies~\cite{EmpiricalHessian, GradSubspace, Goldilock}. This could also be observed from the Hessian density as shown in Fig.~\ref{fig:hessian-stats}. Hence, the average of loss change perturbed by a random Gaussian vector is about $\frac{1}{2}\lambda^2 \sigma^2 tr(H_{\theta})$, which is commonly positive. The variance is hard to calculate because of $tr(H_{\delta}H_{\delta})$. Instead, we simulate the $\delta\mathcal{L}$ by sampling $100$ groups of $\epsilon$ and plot its distribution in the second row of Fig.~\ref{fig:hessian-stats}. The details can be found in Appendix~\ref{supp-sec:demo-soa}. We set $\sigma=1.0$ and plot distributions under $\lambda \in \{0.001, 0.01\}$. The loss change is almost centered around $0.0$ when $\lambda=0.001$, while it becomes almost positive when $\lambda=0.01$. When plotting the 1D curves, we could only sample a limited number of points, and the steps among each other are much greater than $0.001$. For example, if we plot 41 points in the range of $[-1.0, 1.0]$, then the step is 0.05. This step size may commonly lead to loss increment, which explains the GMI phenomenon. Due to formula similarity, MLI could also be explained by this well.

\begin{figure}[tb]
	\centering
	\includegraphics[width=\linewidth]{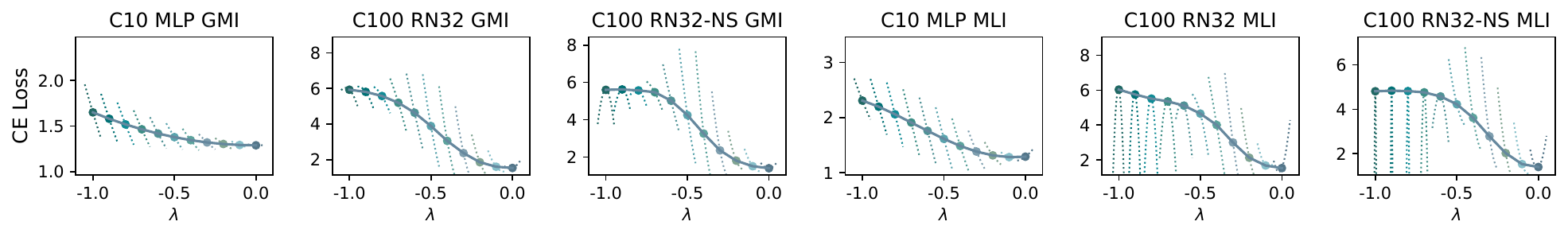}
	\caption{The second-order approximation for the interpolation of $\theta_f + \lambda \epsilon$ with $\lambda \in [-1, 0]$. The first three plots use $\epsilon$ as Gaussian noise (i.e., GMI), while the last ones set $\epsilon=\theta_f-\theta_0$ (i.e., MLI).} \label{fig:theory-approx}
\end{figure}

Finally, we take 11 discrete points to plot the GMI and MLI curves as in Fig.~\ref{fig:theory-approx}. For each discrete point, we plot the second-order approximation curve under a neighbor range of $[-0.05, 0.05]$. The MLI curves look similar to the GMI curves, which verifies that the MLI phenomenon could be explained by perturbing the final model with $\epsilon=\theta_f-\theta_0$. Second, for most discrete points, the s.o.a. shows a decreasing trend when the x-axis becomes larger. It is worth noting that the s.o.a. does not work well when $\lambda=-1$, where the interpolated model is close to the initialization or random noise. The Hessian eigenvalue distribution of the initialization has a slightly large proportion of negative values (Fig.~\ref{fig:hessian-stats}), which also implies that the chaotic loss surface in Fig.~\ref{fig:gaussian} and Fig.~\ref{fig:descent-curves} is more likely to appear on the model worse than the random guess.

\section{Related Works} \label{sec:relate-work}
\noindent \textbf{Low-Dimensional Visualization of Loss Landscape}.
\cite{Goodfellow} plots 1D curves between the initialized and converged model, and between two independently converged models, presenting the interesting monotonic linear interpolation (MLI) phenomenon~\cite{RevisistMLI, AnalyzingMLI, PlateauMLI} and the barrier in linear mode connectivity (LMC)~\cite{OptimizeMC, ModeVolume}. Amazingly, the linear interpolation of two independent solutions only crosses one loss peak, and they could be connected by a slightly complex curve with low losses~\cite{NoBarrier, FGE}. \cite{im2016empirical} empirically shows the surface between solutions found by different optimizers. \cite{VisualizingLandscape} provides visualization of loss landscape with or without skip connections. \cite{VisualBERT} plots 1D curves and 2D surfaces for fine-tuned BERT~\cite{BERT}. Some works also show the guesses about the loss surface of DNNs by plotting demo functions~\cite{Saddle, TheRole, Goldilock}. For example, \cite{Saddle} plots some saddle surfaces and gutter structures by simple 2D functions such as $z=(x^2 + y^2 - 1)^2$.

\noindent \textbf{Global View of Loss Landscape}.
The saddle points may challenge the optimization process of high-dimensional non-convex DNNs~\cite{Saddle}. \cite{MultiLayer} analyzes the global landscape of multi-layer DNNs by replacing the ReLU activation with some assumptions. \cite{NoPoorMinima} proves that deep linear networks have no poor local minima, and \cite{NoBadValley} searches for a special category of DNNs with no bad minima. \cite{Goldilock} proposes a notion of Goldilocks zone to show the effectiveness of proper initialization methods. \cite{LargeScaleLandscape} proposes a toy loss landscape model named n-wedges to present surprising and counter-intuitive properties of DNNs in a more explainable way. Exploring the Hessian matrix of DNNs also explains some interesting properties of DNNs~\cite{EmpiricalHessian, HessianEigenspectrum}. For example, \cite{EigenvaluesBeyond} points out that the Hessian eigenvalue distribution composes the bulk part and the positive outliers, and the number of the latter may be the number of classes. This makes the gradient during optimization lie in a small tiny subspace~\cite{GradSubspace}.

\noindent \textbf{Applications of Studying Loss Landscape}. Exploring the loss landscape could provide some helpful insights for practical applications. The valley flatness or sharpness around a converged model may reflect the generalization performance~\cite{FlatMinima, LargeBatch, ExploreGeneral, Fantastic}, while later works debate against their relation~\cite{Sharp, Modern}. Motivated by the flat minima, some advanced optimization methods are proposed~\cite{Penalizing, Asam, SAM, EntropySGD}. Mitigating the loss barrier between two independent models is studied for better model fusion~\cite{Wagging, OTFusion, GitReBasin, TheRole}, which has also been applied to federated learning~\cite{FedAvg, FedPAN, FedMA}. The asymmetric valley~\cite{AsymmetryValley} explains the success of stochastic weight averaging~\cite{SWA}. Fusing multiple model soups fine-tuned from the same pre-trained model could lead to a better loss area~\cite{WhatIs, ModelSoups}. 

\section{Limitations and Future Works} \label{sec:lim-future}
This paper does not strictly prove how complex the 1-D loss curves of DNNs can be, and it is still an open problem to answer. Additionally, the mined vvv-basin curves are not as perfect as we expected. If an arbitrarily complex 1D curve can be mined, then the global loss surface of the DNNs is more complex. But if we can only mine finite complex 1D curves, then the global loss surface may be either finite complex or may still be complicated by high-dimensional combinations. This is also an open problem to verify. Mining and visualizing more complex types of local geometry structures for various types of DNNs are future works.

\section{Conclusion}
We systematically mine and plot several types of 1D curves embedded in the global landscape of DNNs. Various types of 2D surfaces are further mined from the basis of 1D perturbation directions. Theoretical analysis from the view of Hessian properties explains the GMI and MLI phenomenon. 

\newpage

\appendix
\section{Experimental Details} \label{supp-sec:exper-detail}
This section reports the details about the DNNs and datasets utilized in this paper. Then, the training details and figure details are presented.

\subsection{Datasets, DNNs, and Training Details} \label{supp-sec:exper-data}
An MLP with two layers is investigated on CIFAR-10~\cite{cifar}. We flatten the images in CIFAR-10 to vectors with the size of $32\times 32 \times 3=3072$ as inputs, and the hidden size of the MLP is 128. Then, a classifier layer is used to output 10 classes. ResNet~\cite{ResNet} with 32, 56, and 110 layers are trained on CIFAR-10 and CIFAR-100~\cite{cifar}, and the versions with no-skip connections are also trained. A pre-trained MobileNet-V2~\cite{MobileNetV2} is fine-tuned on CUB~\cite{CUB}. These DNNs are trained for $E=200$ epochs, and the intermediate checkpoints such as $E=10, 50, 100$ are also saved for experimental studies. The learning rate is 0.1 for MLP and ResNet, and a smaller one is 0.01 for pre-trained MobileNet-V2. The batch size is 128 and the weight decay is $5\times 10^{-4}$. Additionally, the pre-trained ResNeXt101 (RNX101)~\cite{ResNeXt} could be directly downloaded from \texttt{PyTorch} and verified on ImageNet~\cite{ImageNet}.

\subsection{Processing of BN Layers} \label{supp-sec:bn}
Some utilized networks contain the BN layers~\cite{bn}, which contain two types of parameters. The first type contains ``BN.weight'' and ``BN.bias'', which are trainable during the loss backward pass. The other ones contain running statistics such as ``BN.running\_mean'' and ``BN.running\_var'', which calculate the mean and variance of hidden representations channel by channel in the forward pass. When perturbing $\theta$ by the noise direction $\epsilon$, the running statistics are not considered. Hence, the interpolated model $\theta + \lambda \epsilon$ may have inconsistent running statistics, which should be updated by additionally taking a forward pass of the data to re-calculate them. The processing of this manner is referred to as ``UpBN'' by default. If we do not additionally update the running statistics and directly use the ones in $\theta$ for $\theta + \lambda \epsilon$, we name this manner as ``No UpBN''.

\subsection{Figure Details and More Plots} \label{supp-sec:fig}
We then respectively present some details when plotting the figures of Fig.~\ref{fig:gaussian} and Fig.~\ref{fig:gaussian-noskip}. For each Gaussian noise direction, we scale its norm to the same as the perturbed model by $\epsilon \leftarrow ||\theta|| * \epsilon / ||\epsilon||$. The range of $\lambda$ is $[-1, 1]$ by default, i.e., $s=1.0$. For each 1D curve, we calculate the number of stationary points. The stationary point means that its $y$-axis value is both larger or smaller than that of its left and right point, i.e., satisfying $(y_t - y_{t+1})*(y_t - y_{t-1}) > 0$. We do not consider the two endpoints in the 1D curve. Perfect v-basin curves have 1 stationary point, and perfect w-basin/w-peak curves have 3 stationary points. The number of stationary points that one 1D curve has could reflect its smoothness. As empirically pointed out by this paper, it is hard to find and plot 1D loss curves of DNNs that have more than 5 stationary points.

Various conditions are considered to verify the Gaussian Monotonic Increasing (GMI) phenomenon, including the number of training epochs ($E \in \{0, 10, 50, 100, 200\}$), the normalization ways of the Gaussian noise (``Norm'', ``LayNorm'' and ``FilNorm''), the processing of BN statistics (``UpBN'' and ``No UpBN''), and the view scale ($s \in \{0.1, 1.0, 4.0\}$).

``Norm'' means that we may normalize the perturbation direction to have the same norm of the perturbed model, i.e., $\epsilon \leftarrow ||\theta||\frac{\epsilon}{||\epsilon||}$. This is utilized in~\cite{im2016empirical}. The filter normalization (``FilNorm'') is proposed by~\cite{VisualizingLandscape}, which normalizes each filter in the noise direction to have the same norm of corresponding filters in the perturbed model. We also utilize the ``LayNorm'' which normalizes each layer in the noise direction to have the same norm of corresponding layers in the perturbed model. ``UpBN'' means that we re-calculate the running statistics in the BN layer (i.e., the running mean and variance) by taking an additional forward pass. ``No UpBN'' means that we keep the BN statistics as the ones in the perturbed model. The view scale means that we plot 1D curves or 2D surfaces in the range of $[-s, s]$ instead of $[-1, 1]$.

\begin{figure}[tb]
	\centering
	\includegraphics[width=\linewidth]{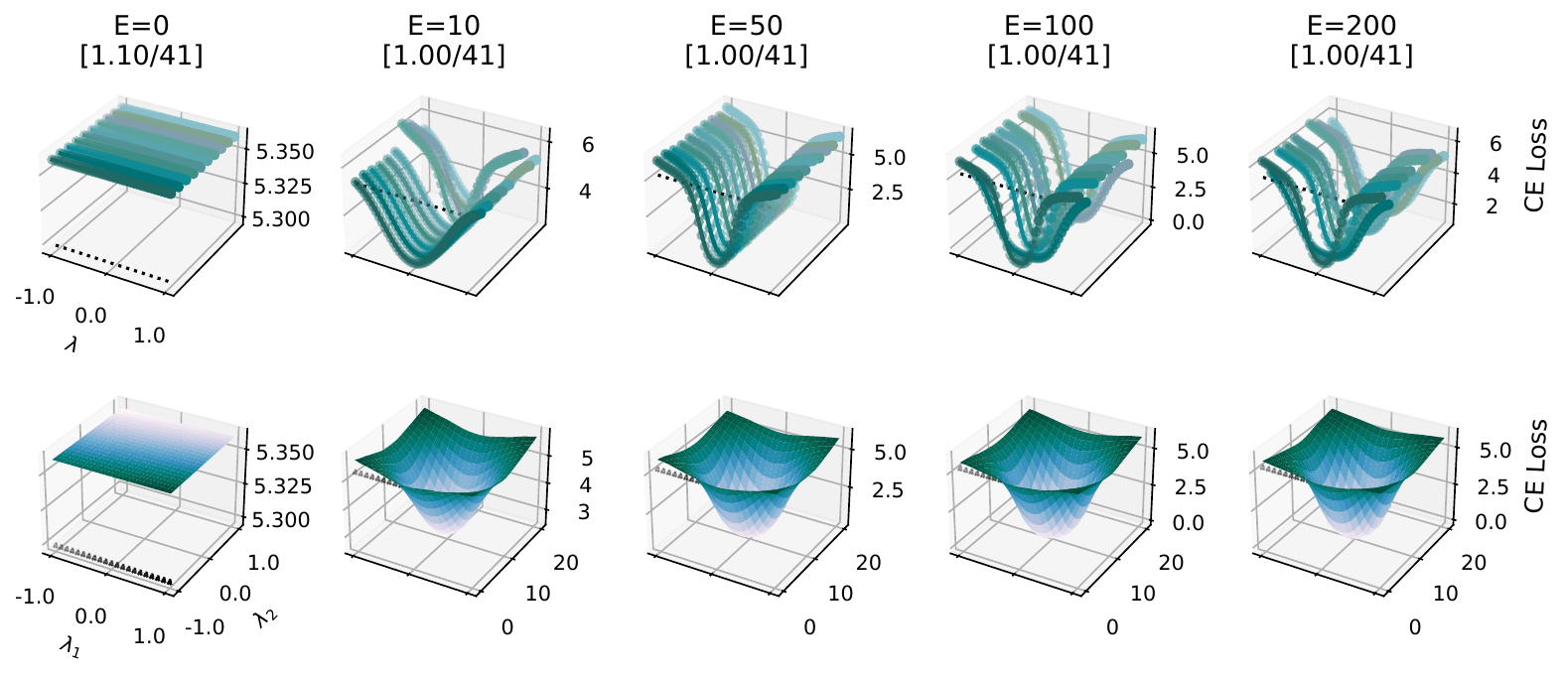}
	\caption{The GMI phenomenon under various checkpoints of MV2 on CUB.} \label{fig:supp-ckpt}
\end{figure}

\begin{figure}[tb]
	\centering
	\includegraphics[width=\linewidth]{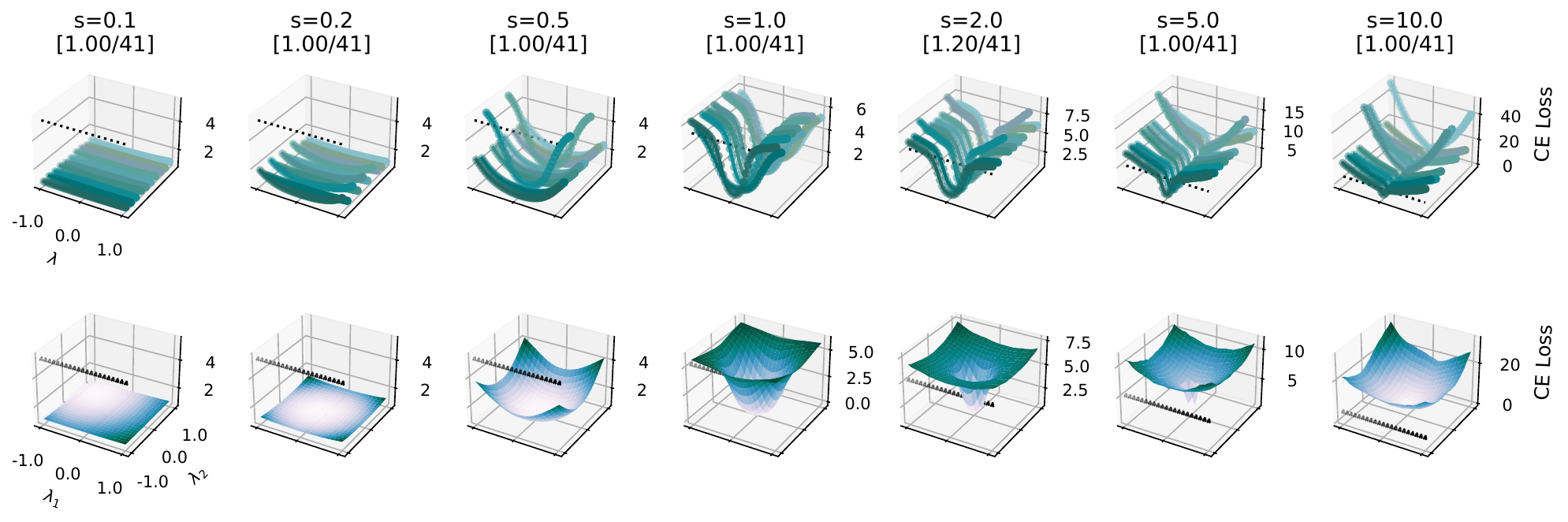}
	\caption{The GMI phenomenon under various view scales of MV2 on CUB.} \label{fig:supp-fine}
\end{figure}

\begin{figure}[t]
	\centering
	\includegraphics[width=\linewidth]{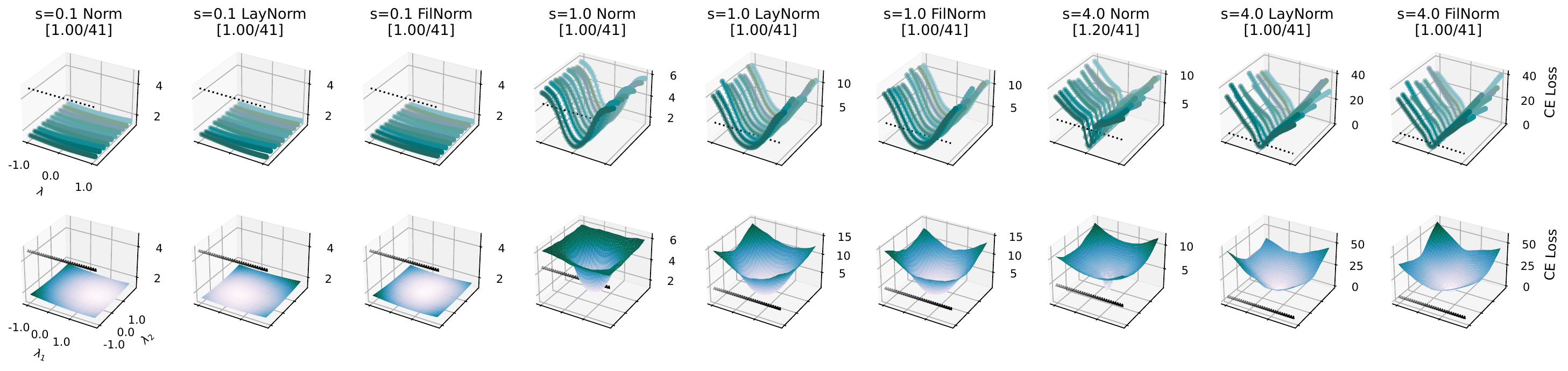}
	\caption{The GMI phenomenon under various view scales and various normalization ways of RN32 on C100.} \label{fig:supp-filter}
\end{figure}

With these conditions, we additionally propose several groups of plots to verify the GMI phenomenon. Fig.~\ref{fig:supp-ckpt} shows the conditions under different checkpoints of MV2 on CUB. Fig.~\ref{fig:supp-fine} shows the conditions under various view scales of $s \in \{0.1, 0.2, 0.5, 1.0, 2.0, 5.0, 10.0\}$ on CUB with MV2. Fig.~\ref{fig:supp-filter} shows the conditions under various view scales and various normalization ways on C100 with RN32. All plots show obvious GMI phenomenon.

\subsection{A Summary of Mined 1D Curves and Corresponding Perturbation Directions} \label{supp-sec:summary}
In this paper, we gradually visualize and mine several types of 1D curves by category. The following lists the summary of these types of curves.
\begin{itemize}
    \item \textbf{v-basin curves}: amazingly, random Gaussian directions could almost lead to v-basin curves (Fig.~\ref{fig:gaussian}), and it is hard to generate other types of curves such as v-side ones (Fig.~\ref{fig:descent-steps}). The Gaussian perturbation is abbreviated as ``Gau.'' in Fig.~\ref{fig:combine-2d}. Aside from the Gaussian direction, the eigenvectors corresponding to the positive eigenvalues of the Hessian matrix also display v-basin curves (Fig.~\ref{fig:hessian-1d}), which are abbreviated as ``P.E.x'' in Fig.~\ref{fig:combine-2d}. ``x'' ranges from 1 to 10.
    \item \textbf{v-side curves}: the negative gradient and the direction to the subsequent checkpoints are intuitive descent directions (Fig.~\ref{fig:descent-curves}). The negative gradient is abbreviated as ``Neg. Grad.''. The direction to subsequent checkpoints is abbreviated as ``$E=200$'' (or other checkpoints) in Fig.~\ref{fig:combine-2d}.
    \item \textbf{w-basin curves}: the direction to an independent checkpoint is an intuitive direction for plotting w-basin curves (Fig.~\ref{fig:w-curves}), which is abbreviated as ``$E'=200$'' (or other checkpoints) in Fig.~\ref{fig:combine-2d}.
    \item \textbf{w-peak curves}: the eigenvectors corresponding to the negative eigenvalues of the Hessian matrix lead to the w-peak curves (Fig.~\ref{fig:hessian-1d}), which are abbreviated as ``N.E.x'' in Fig.~\ref{fig:combine-2d}. ``x'' ranges from 1 to 10. The w-peak curves could also be plotted by the directions mined by Algo.~\ref{algo:mine}, which are abbreviated by ``Mine'' in Fig.~\ref{fig:combine-2d}.
    \item \textbf{vvv-basin curves}: the directions mined by Algo.~\ref{algo:mineplus} may lead to vvv-basin curves. However, the vvv-basin pattern is not obvious, and we do not use it to plot 2D surfaces in Fig.~\ref{fig:combine-2d}.
\end{itemize}

Most of the 1D directions are orthogonal to each other, and hence, we could plot the 2D surfaces without further processing.

\section{Demo Code} \label{supp-sec:demo-code}
This section provides some demo codes for quickly reproducing some experimental studies, including the calculation of the Hessian eigenvectors by \texttt{sparse.linalg.eigsh}, the calculation of the eigenvalue density of the Hessian by \texttt{PyHessian}, and the second approximation as in Eq.~\ref{eq:approx}.

The core part of these codes is the ``Jacobian Vector Product'' function that returns $H_{\theta}v$ for a given vector $v$. This is usually implemented by sequentially taking two passes of backward process. Specifically, the first backward pass of the loss could obtain the gradient $g=\nabla_{\theta}\mathcal{L}$. And we calculate $g^Tv$ as the loss and backward again, which leads to $\nabla_{\theta}(g^Tv) = (\nabla_{\theta} g^T)v = Hv$. This just returns the product of the Hessian $H$ and the vector $v$. With this trick, we do not need to completely compute the whole Hessian matrix itself, which is hard to calculate on a limited computation and storage resource.

\subsection{Calculating the Hessian eigenvectors by \texttt{sparse.linalg.eigsh}} \label{supp-sec:demo-eigsh}
If we aim to obtain the largest Hessian eigenvalue and corresponding eigenvector, the power iteration method could be utilized. Specifically, given a random vector $v$, we could keep calculating the ``Jacobian Vector Product'' by $v \leftarrow \frac{Hv}{||Hv||}$. After convergence, we could obtain the largest eigenvalue and corresponding eigenvector. However, calculating the smallest eigenvalue and corresponding eigenvector is slightly complex. We utilize the package of \texttt{sparse.linalg.eigsh} to accomplish this goal. \texttt{eigsh} accepts as the input a square operator representing the operation $Hv$, where $H$ is real symmetric or complex Hermitian. This condition is satisfied by the Hessian matrix and the trick of ``Jacobian Vector Product''. The demo code is listed in Code, which also utilizes the \texttt{torch.autograd} and \texttt{LinearOperator}. We omit the introduction of these functions and these could be found on the web easily. The function parameter ``which'' determines the types of eigenpairs. If ``which'' is set as ``LA''/``SA'', the function calculates the largest/smallest algebraic eigenvalues and corresponding eigenvectors. We calculate the largest top $k=10$ eigenpairs and smallest $k$ eigenpairs and denote the eigenvectors as ``P.E.x'' and ``N.E.x'' respectively. The value of ``x'' ranges from 1 to 10, and a smaller ``x'' refers to that the corresponding eigenvalue has a larger absolute value.

\subsection{Calculating the eigenvalue density of the Hessian by \texttt{PyHessian}} \label{supp-sec:demo-pyhessian}
\texttt{PyHessian}~\cite{PyHessian} is a package that calculates the statistics of the Hessian matrix for DNNs, which includes: (1) the most top-k largest eigenvalues and corresponding eigenvectors that utilize the power iteration method; (2) the trace of the Hessian matrix that could be estimated by $\mathbb{E}_{\epsilon}[\epsilon^TH\epsilon]$ where $\epsilon$ is element-wisely sampled from $\mathcal{N}(0, 1)$; (3) the eigenvalue density of eigenvalues approximated by the algorithm of Stochastic Lanczos Quadrature (SLQ)~\cite{SLQ}. We download the source codes of \texttt{PyHessian} and utilize the ``\texttt{PyHessian}'' class and the ``\texttt{density\_generate}'' function to plot the eigenvalue density. 


\subsection{Simulating the Second Approximation as in Eq.~\ref{eq:approx}} \label{supp-sec:demo-soa}
As shown in Eq.~\ref{eq:approx}, the second-order approximation formulation is $\delta\mathcal{L} = \mathcal{L}(\theta + \lambda\epsilon) - \mathcal{L}(\theta) \approx \lambda \epsilon^T g_{\theta} + \frac{1}{2}\lambda^2 \epsilon^T H_{\theta} \epsilon$. This could be viewed as a quadratic function of $\lambda$, i.e., $\frac{1}{2}a\lambda^2 + b\lambda$ with $a=\epsilon^T H_{\theta} \epsilon$ and $b=\epsilon^T g_{\theta}$. Hence, given a model point $\theta$, we could calculate $a$ and $b$ by the trick of ``Jacobian Vector Product'' again. For each given $\lambda$, we could sample multiple groups of $\epsilon$ and then plot the distribution of $\delta\mathcal{L}$. 

To approximate the 1D curves in Fig.~\ref{fig:theory-approx}, we calculate the $a=\epsilon^T H_{\theta+\lambda\epsilon} \epsilon$ and $b=\epsilon^T g_{\theta+\lambda\epsilon}$ for the model point $\theta+\lambda\epsilon$, and then plot the quadratic curve by $\frac{1}{2}a(x-\lambda)^2 + b(x - \lambda)$ with $x\in [\lambda-0.05, \lambda+0.05]$.




\bibliographystyle{elsarticle-num}
\bibliography{landscape.bib}
\end{document}